# Retrosynthetic reaction prediction using neural sequence-to-sequence models

Bowen Liu[1], Bharath Ramsundar[2], Prasad Kawthekar[2], Jade Shi[1], Joseph Gomes[1], Quang Luu Nguyen[1], Stephen Ho[1], Jack Sloane[1], Paul Wender[1,3], Vijay Pande[1,2,4]

[1] Department of Chemistry, Stanford University, Stanford, CA 94305, USA

[2] Department of Computer Science, Stanford University, Stanford, CA 94305, USA

[3] Department of Chemical and Systems Biology, Stanford University, Stanford, CA 94305, USA

[4] Department of Structural Biology, Stanford University, Stanford, CA 94305, USA

Email: pande@stanford.edu

**Abstract:**

We describe a fully data driven model that learns to perform a retrosynthetic reaction prediction task, which is treated as a sequence-to-sequence mapping problem. The end-to-end trained model has an encoder-decoder architecture that consists of two recurrent neural networks, which has previously shown great success in solving other sequence-to-sequence prediction tasks such as machine translation. The model is trained on 50,000 experimental reaction examples from the United States patent literature, which span 10 broad reaction types that are commonly used by medicinal chemists. We find that our model performs comparably with a rule-based expert system baseline model, and also overcomes certain limitations associated with rule-based expert systems and with any machine learning approach that contains a rule-based expert system component. Our model provides an important first step towards solving the challenging problem of computational retrosynthetic analysis.

**Introduction**

Organic synthesis is a critical discipline that directly brings scientific and societal benefits by enabling access to poorly available molecules and, more significantly, to new molecules that have never been studied before. This accessibility fundamentally enables other fields of research such as materials science, environmental science, and drug discovery. Retrosynthetic analysis is a technique widely-used by organic chemists to design synthetic routes to "target" molecules, where the target is recursively transformed into simpler precursor molecules until commercially available "starting" molecules are identified.[1–3] It encompasses two related tasks. The first task, reaction prediction, involves predicting how a set of reactants will react to form products. The second task involves planning the optimal series of reaction prediction steps to recursively deconvolute the target molecule into simple or commercially available precursor molecules in a way that minimizes steps, cost, time, and waste.[4,5]

Computational retrosynthetic analysis tools can potentially greatly assist chemists in designing synthetic routes to novel molecules, and would have many applications in drug discovery, medicinal chemistry, materials science and natural product synthesis. Since the 1960s, chemists have recognized the promise of modern computing in assisting organic synthesis analyses. However, although various algorithms have been developed over the years, their widespread acceptance by mainstream chemists has lagged.[6] In part this is due to these approaches being applicable to only relatively simple target molecules for which expert chemists could readily deduce synthetic plans without assistance.[7] The first class of algorithms uses reaction rules that are either manually encoded by human experts or automatically derived from a reaction database.[8–23] The key drawback of such rule-based expert systems is that they generally cannot make accurate predictions outside of their knowledge base. As a result, these systems perform poorly when generalizing to new target structures and reaction types. The second class of algorithms uses principles of physical chemistry to predict energy barriers of a reaction based on first principles.[24–30] Although this approach can generalize to novel molecules and reaction types, currently such calculations are often prohibitively expensive to perform for full synthetic planning problems.

The third class of algorithms is based on machine learning techniques, which attempts to address the shortcomings of the rule-based and the physical chemistry approaches by making predictions that generalize better than those of rule-based approaches at a computational cost that is much less than those of physical chemistry approaches.[31–34] More recently, deep learning techniques have been applied to the reaction prediction task. The typical deep learning approach combines a rule-based expert system with a feedforward neural network (NN) component that performs candidate ranking. The NN either ranks the applicability of each rule in the knowledge base to a given example,[35,36] or ranks the likelihood of each predicted product obtained by applying all the rules in the knowledge base to a given example.[37]

However, these types of deep learning approaches are fundamentally dependent on the rule-based expert system component and thus inherit some of its major limitations. In particular, these approaches have issues with making accurate predictions outside of the rule-based knowledge base. Additionally, there is a tradeoff between defining very general rules that result in a lot of noise, and defining very specific rules that are only applicable to a limited set of reactions with very specific reactants and products.[35] The reaction rules are necessarily inadequate representations of the underlying chemistry because they focus on the local molecular environment of the reaction centers only. Furthermore, the rule-based expert system components used by these deep learning approaches do not fully account for stereochemistry; noticeably, none of their reported reaction examples contain molecules with stereocenters.

An alternative deep learning approach that eliminates the rule-based expert system component would overcome these limitations. One way to view the reaction prediction task is to cast it as a sequence-to-sequence prediction problem, where the objective is to map a text sequence that represents the reactants to a text sequence that represents the product, or vice versa. Although molecules are usually represented as 2D or 3D graphs, they can also be equivalently expressed as text sequences in line notation format, such as simplified molecular-input line-entry system (SMILES)[38] or International Chemical Identifier (InChI).[39] The text representation of molecules in various chemoinformatic applications have been explored previously.[40–44] Recently, Nam and Kim[45] described a neural sequence-to-sequence (seq2seq) model for the forward reaction prediction task. The model was trained end-to-end on a combination of artificially generated reactions and experimental reactions from an open source patent dataset.[46] Given an input SMILES that represents the reactants, the model directly outputs a SMILES that represents the predicted products.

Here, we attempt to solve the retrosynthetic reaction prediction task, which presents additional challenges compared to the forward reaction prediction task, because the input in the retrosynthetic reaction prediction task contains less information and there are many more possible outputs. For forward reaction prediction, the starting materials significantly constrain the reaction types that are possible and limits the number of possible products. On the other hand, for retrosynthetic reaction prediction, there are usually multiple possible ways to disconnect the target molecule, via many different reaction types, to produce a large number of possible starting materials. Indeed, every bond in the target molecule represents a possible retrosynthetic disconnection. Figure 1 shows a synthetic scheme for phenylalanine that illustrates the asymmetry in the input constraints for the forward and retrosynthetic reaction prediction tasks. In the forward direction, each of the three different sets of starting materials and reaction conditions will result in phenylalanine as the single major product. Conversely, in the retrosynthetic direction, the phenylalanine can be disconnected into three different sets of starting materials.

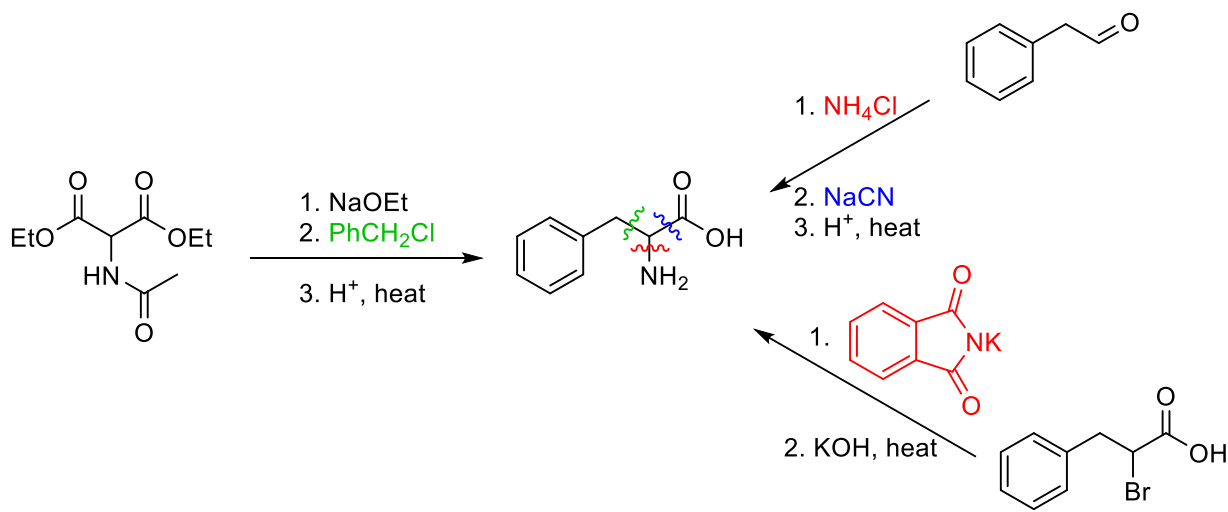

Figure 1: Phenylalanine synthetic scheme

In this work, we describe our initial studies directed at solving the challenging problem of computational retrosynthetic analysis. In particular, we develop a fully data driven seq2seq model that learns to perform the retrosynthetic reaction prediction subtask. For a given target molecule and a specified reaction type, the model predicts the most likely reactants that can react in the specified reaction type to produce the target molecule. The seq2seq model is trained end-to-end on a subset of experimental reactions with labelled reaction types[47] from an open source patent database.[46] We show that the trained seq2seq model performs comparably with a rule-based expert system baseline model on the relatively simple chemistry found in the patent dataset.

**Approach**

*Problem definition*

Concretely, the retrosynthetic reaction prediction task is shown in Figure 2. Given an input SMILES that represents the target molecule and a specified reaction type, the model predicts the output SMILES which represents the likely reactants that can react in the specified reaction type to form the target molecule.

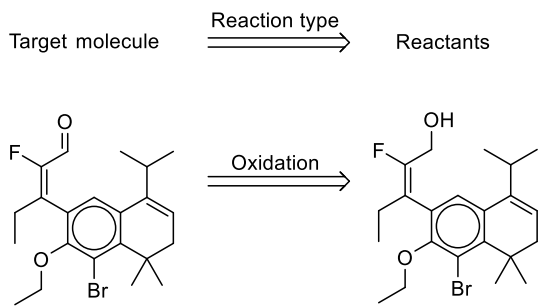

*Figure 2: Retrosynthetic reaction prediction task and an example of a possible retrosynthetic disconnection for a target molecule*

*Data preparation*

We use a filtered patent dataset, derived from an open source patent database,[46] which contains 50,000 atom-mapped reactions that have been classified into 10 broad reaction types.[47] This filtered patent dataset was originally constructed to represent the typical reaction types found in the medicinal chemist's toolkit. The reaction examples are further preprocessed to eliminate all reagents in order to only contain reactants and products,[47] and then canonicalized. We additionally process this dataset so that each reaction example contains a single product by splitting any reactions with multiple products into multiple single product reactions that contain the original reactants. Any resulting reaction examples with trivial products such as inorganic ions and solvent molecules are removed. Table 1 shows the distribution of the 10 reaction classes in the final processed dataset. Finally, the dataset was split into training, validation and test datasets (8:1:1).

*Table 1: Distribution of major reaction classes within the processed reaction dataset*

| reaction class | reaction name | # examples |
|---:|---|---:|
| 1 | Heteroatom alkylation and arylation | 15204 |
| 2 | Acylation and related processes | 11972 |
| 3 | C-C bond formation | 5667 |
| 4 | Heterocycle formation | 909 |
| 5 | Protections | 672 |
| 6 | Deprotections | 8405 |
| 7 | Reductions | 4642 |
| 8 | Oxidations | 822 |
| 9 | Functional group interconversion (FGI) | 1858 |
| 10 | Functional group addition (FGA) | 231 |

*Model*

*Seq2seq model*

Neural sequence-to-sequence (seq2seq) models map one sequence to another, and have recently shown state of the art performance in many tasks such as machine translation.[48,49] It is based on an encoder-decoder architecture that consists of two recurrent neural networks (RNN), and can include an attention mechanism that aligns the target tokens with the source tokens.[48] Figure 3 shows a simple seq2seq encoder-decoder architecture for our retrosynthetic reaction prediction task.

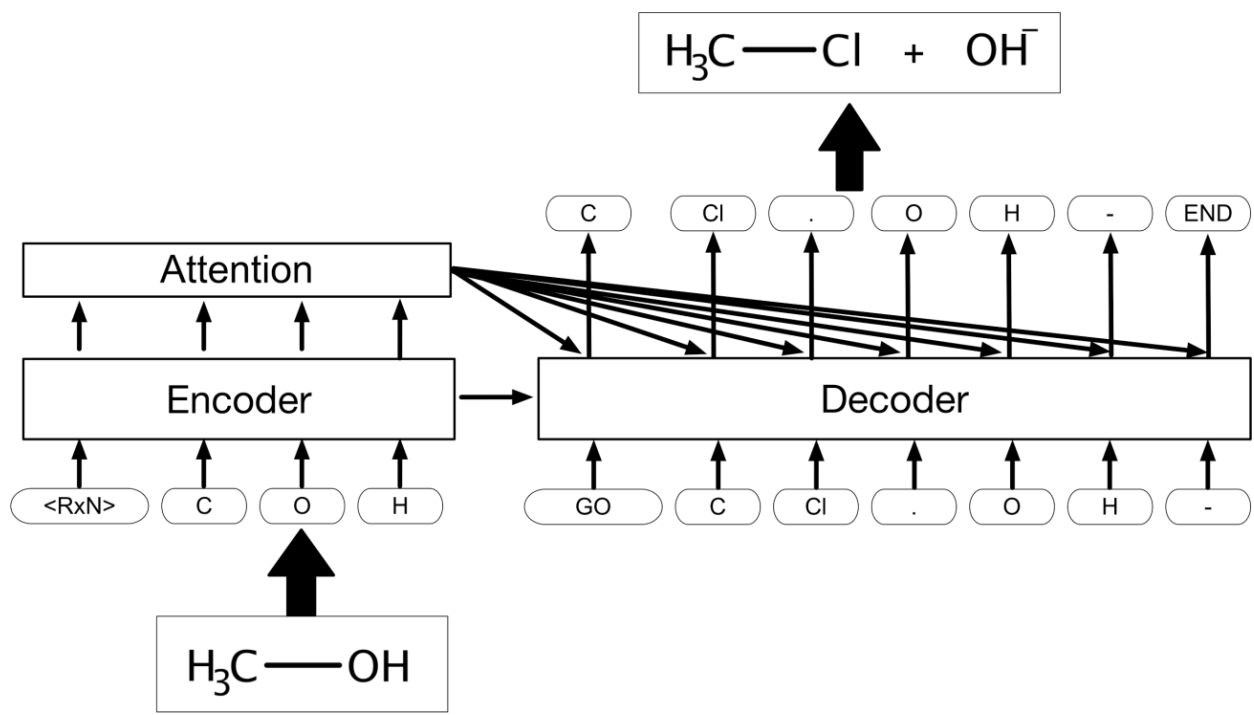

*Figure 3: seq2seq model architecture*

We adapt the open source seq2seq library from Britz et al.[50] for our character-wise seq2seq model. The encoder-decoder architecture consists of Long Short Term Memory (LSTM) cells, which is a variant of RNN cells that more effectively learn long-range dependencies in the sequences.[51] More specifically, the seq2seq model consists of a bidirectional LSTM encoder and a LSTM decoder. Furthermore, an additive attention mechanism is used.[48] The key hyperparameter settings of the seq2seq model are shown in Table S1.

The seq2seq model is trained on the training dataset with reaction atom-mapping removed. Each reaction example is split into a source sequence and target sequence. The source sequence consists of a sequence of characters that is derived from splitting the SMILES that correspond to the product into characters, with a reaction type token prepended to the sequence. The source sequence is reversed prior to feeding into the encoder. The target sequence consists of a sequence of characters that is derived from splitting the SMILES that correspond to the reactants into characters. The seq2seq model is evaluated every 4000 training steps on the validation dataset, and model training is stopped once the evaluation log perplexity starts to increase.

Finally, the trained seq2seq model is evaluated on the test dataset with reaction atom-mapping removed. Each target molecule SMILES example from the test dataset is converted into an input sequence of characters, with a reaction type token prepended to the sequence, and reversed prior to feeding into the encoder. A beam search procedure is used for model inference. Figure 4 depicts a partially completed beam search procedure with a beam width of 5 for an example input. For each source sequence input that represents the target molecule, the top N candidate output sequences ranked by overall sequence log probability at each time step during decoding are retained, where N is the width of the beam. The decoding is stopped once the lengths of the candidate sequences reach the maximum decode length of 140 characters. The candidate sequences that contain an end of sequence character are considered to be complete. These complete candidate sequences represent the reactant sets predicted by the seq2seq model for a particular target molecule, and they are ranked by the overall sequence log probabilities, which consists of the log probabilities of the individual characters in each complete candidate sequence.

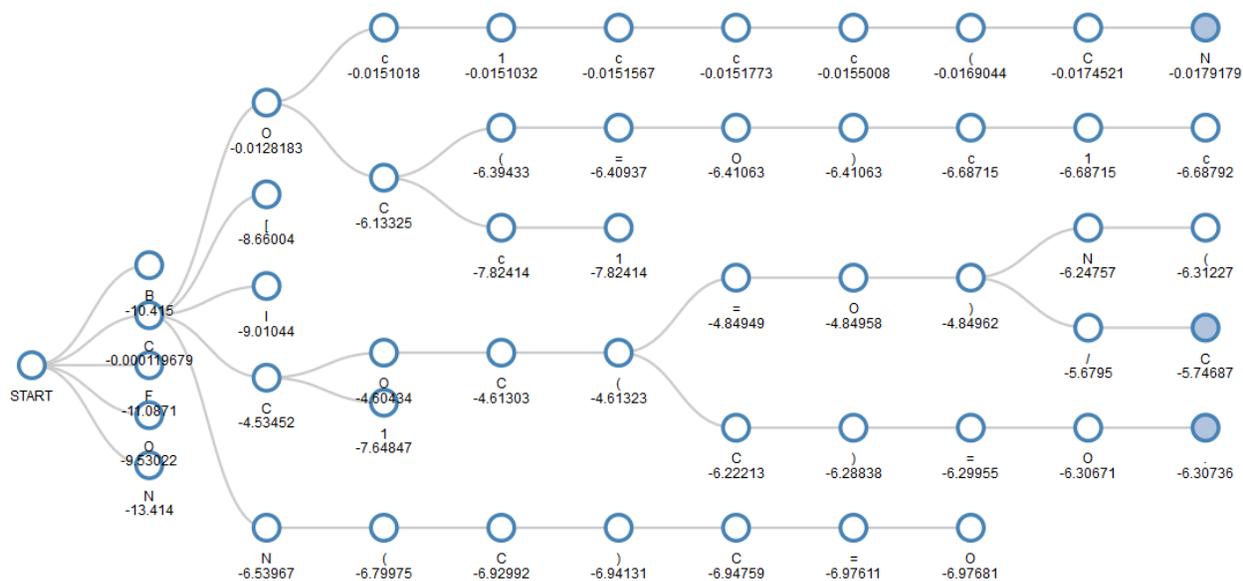

*Figure 4: A partially completed beam search procedure with a beam width of 5 for an example input. Note that only the top 5 candidate sequences are retained at each time step. The visualization was produced using the seq2seq model library from Britz et al.[50]*

*Baseline model*

The baseline model is a rule-based expert system that applies retrosynthetic reaction rules of a specified reaction type to a target molecule to obtain the reactants. The reaction rules are automatically extracted from the training dataset. The rule extraction algorithm is adapted from Coley et al.'s implementation,[37] which was based on the algorithms described by Law et al.[20] and Bogevig et al.[52] For each atom-mapped reaction example in the training dataset, the reaction centers are extracted by identifying changes in connectivity between product atoms and the corresponding reactant atoms. The reaction centers are expanded to include immediately neighboring atoms. A SMARTS string that describes the reaction core pattern is generated for the reactants and product, and combined to form a reaction SMARTS string that represents the retrosynthetic reaction rule. Each reaction rule is labelled with the reaction type of the corresponding reaction example that it was extracted from. Overall, 29462 valid rules were extracted from the training dataset, which represents a rule coverage of 73.1% in the training dataset. A rule is defined to be valid if it is able to regenerate the product from the reactants and the reactants from the product in the reaction example from which the rule is extracted from. After filtering out duplicated rules, we end up with 2861 unique retrosynthetic reaction rules, defined by SMARTS strings.

The rule-based expert system is evaluated on the test dataset. Each target molecule SMILES example from the test dataset is applied by all the rules of the particular reaction type. The resulting top N reactant sets obtained from the successful reaction rules are ranked by the number of occurrences of the corresponding rule of the target reaction class that were observed in the training dataset.

All scripts were written in Python (version 3.5), and RDKit (version 2016.09.04)[53] was used for reaction preprocessing and rule extraction. The seq2seq model was built with TensorFlow (version 1.0.1).[54]

**Results**

*Performance on the test dataset*

Table 2 shows the top-N accuracies of the rule-based expert system baseline and the seq2seq model on the test dataset. The top-N accuracy refers to the percentage of examples where the ground truth reactant set, which is the actual patent literature reported reactant set for the corresponding target molecule in the test dataset, was found within the top N predictions made by the model. By this metric, we observed that the seq2seq model performs comparably to the baseline model. Although the baseline model does not incorporate any NN component to perform candidate ranking, the performance of the baseline model becomes less sensitive to candidate ranking as N increases. The maximum accuracy of the baseline model, which is the percentage of examples where the ground truth reactant set was found in any of the predictions made by the baseline model, is 69.8%. This represents the maximum possible test accuracy of the baseline model, as well as any deep learning approach that combines a NN component that performs candidate ranking with this rule-based expert system. The reason is that if no reaction rule exists that can produce the ground truth reactant set from the input molecule, then the NN component cannot rank it. The top-50 accuracy of the seq2seq model is higher than this maximum baseline accuracy.

*Table 2: Comparison of top-N accuracies between the baseline and seq2seq models*

| model | top-N accuracy (%) | | | | | |
|---|---|---|---|---|---|---|
| | top-1 | top-3 | top-5 | top-10 | top-20 | top-50 |
| baseline | 34.8 | 53.2 | 59.8 | 65.7 | 69.0 | 69.8 |
| seq2seq | 34.1 | 51.1 | 56.5 | 62.0 | 66.8 | 71.9 |

Some representative examples of correct seq2seq model predictions for each reaction class are shown in Figure 5. The examples are depicted in the retrosynthetic direction.

1: Heteroatom alkylation and arylation

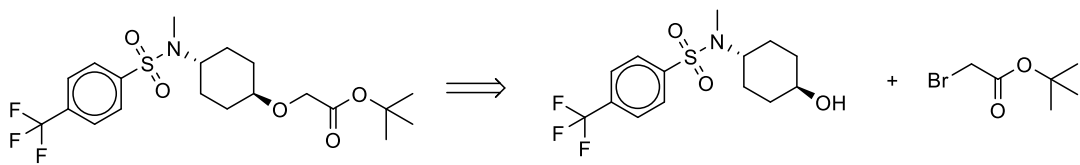

2: Acylation and related processes

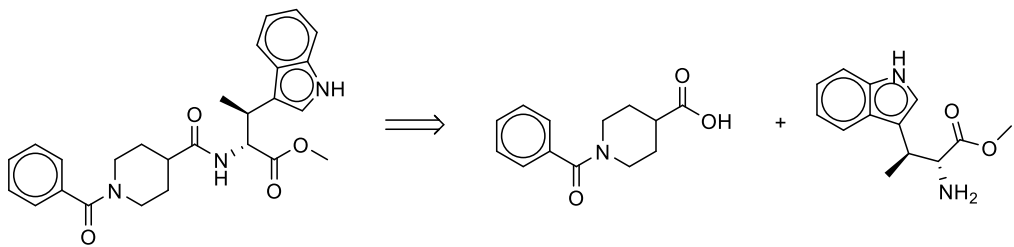

3: C-C bond formation

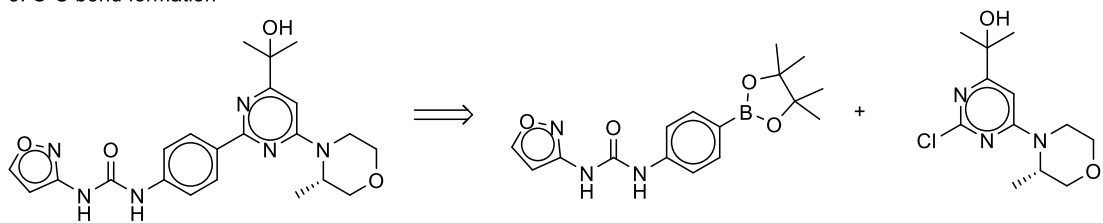

4: Heterocycle formation

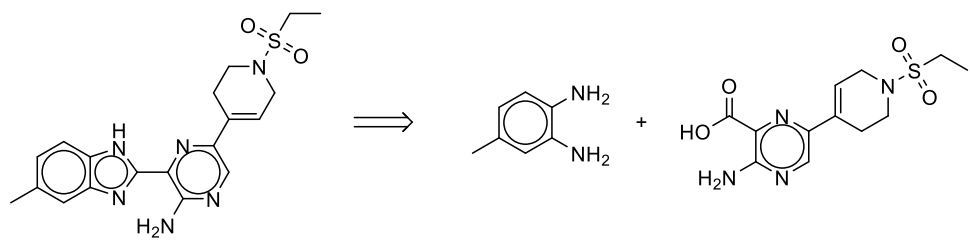

5: Protections

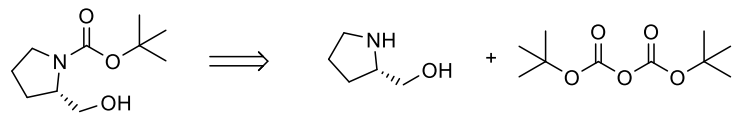

6: Deprotections

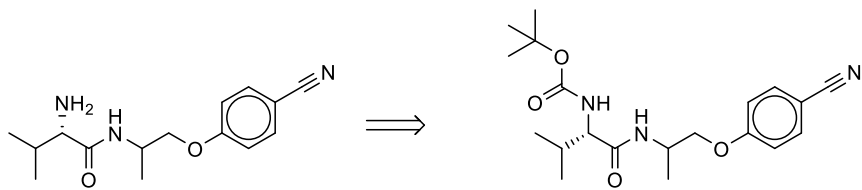

7: Reductions

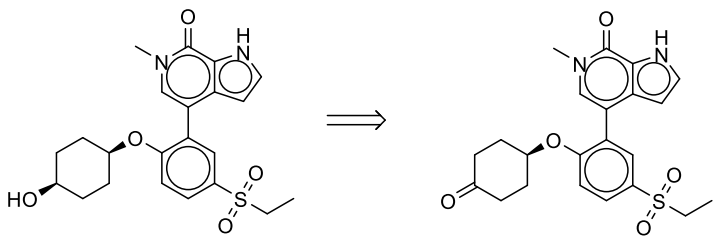

8: Oxidations

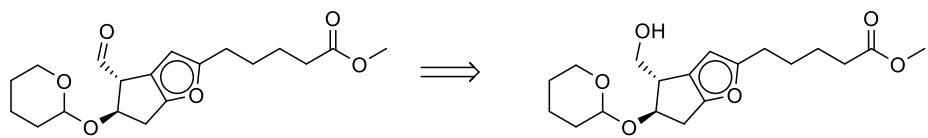

9: Functional group interconversion (FGI)

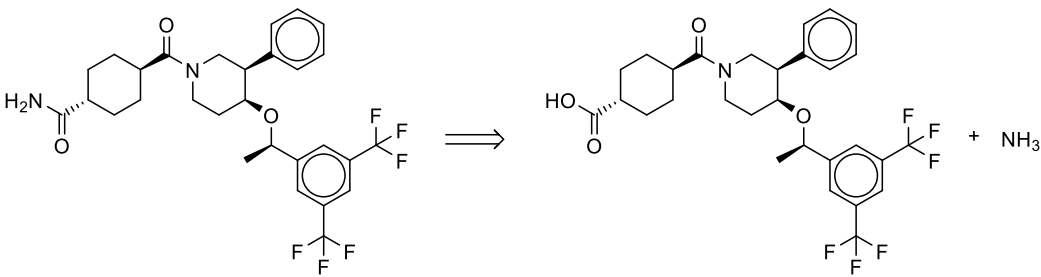

10: Functional group addition (FGA)

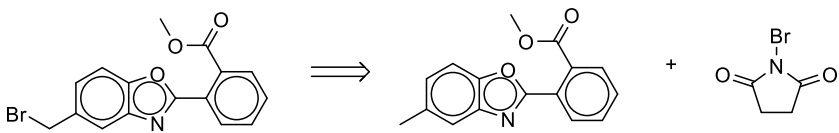

*Figure 5: Representative examples of correct seq2seq model predictions for each reaction class*

The detailed top-10 results for the baseline model and the seq2seq model broken down by the reaction classes is shown in Table 3. The name of each reaction class is shown in Table 1.

*Table 3: Breakdown of the top-10 accuracy of the baseline and seq2seq models by reaction class*

|  | reaction class | | | | | | | | | |
|---|---|---|---|---|---|---|---|---|---|---|
|  | 1 | 2 | 3 | 4 | 5 | 6 | 7 | 8 | 9 | 10 |
| *baseline top-10 accuracy (%)* | 78.6 | 85.7 | 55.3 | 47.3 | 14.9 | 24.7 | 75.9 | 62.5 | 50.5 | 82.6 |
| *seq2seq top-10 accuracy (%)* | 57.6 | 75.2 | 50.4 | 18.7 | 79.1 | 59.2 | 70.5 | 64.6 | 50.5 | 82.6 |
| *# examples* | 1521 | 1197 | 567 | 91 | 67 | 841 | 464 | 82 | 186 | 23 |

The baseline model performs significantly better in reaction class 1 (heteroatom alkylation and arylation) and reaction class 2 (acylation and related processes). The common feature of these reaction classes is that the reactions are possible with many possible functional groups at the reaction site. For example, in Figure 5, the acylation reaction between a carboxylic acid and an amine would also be possible with another carbonyl compound that has a suitable leaving group, such as an acyl chloride. Additionally, the target molecules in the dataset for these reaction classes often have multiple possible reaction sites. The knowledge base in the baseline model contains reaction rules for each of the possible functional groups that were present in the reaction examples from the training dataset. Therefore, the baseline model is able to easily enumerate reactant sets that span most of the possible functional group and reaction site combinations. On the other hand, the seq2seq model currently predicts only a few valid reactant sets that contain different possible functional group and reaction site combinations. As a result, the particular ground truth reactant set is more likely to be found in the predicted reactant sets from the baseline model.

The baseline model also performs significantly better in reaction class 4 (heterocycle formation). The key feature of this reaction class is the formation of cyclic and aromatic structures, which results in a large difference between the reactant set SMILES string and the target molecule SMILES string. Also, there is a relatively small number of reaction examples in the training dataset for this reaction class. Overall, these two factors cause the seq2seq model to make a lot of grammatical mistakes in the SMILES predictions.

The seq2seq model performs significantly better in reaction class 5 (protections) and reaction class 6 (deprotections). The common feature of reaction classes 5 and 6 is that the reactants have large leaving groups that are not included in the product side. As a result, the very general rules in the baseline model, which only contain the immediate neighborhood of the reaction centers, do not capture the identities of the leaving groups. Conversely, the seq2seq model captures the global molecular environment of all the reaction species and is able to predict the leaving groups correctly.

*Error analysis of the seq2seq model*

The seq2seq model makes three kinds of prediction errors:

1. The predicted reactant SMILES is grammatically invalid. This is a result of the SMILES text representation of the molecules, which is fragile because single character alterations can completely invalidate the SMILES. Since the seq2seq decoder does not explicitly understand the grammar that underlie the SMILES representation, and also formulates predictions one character at a time, it is likely that some of the predicted reactant SMILES are invalid. Figure 6 shows examples of this type of error.

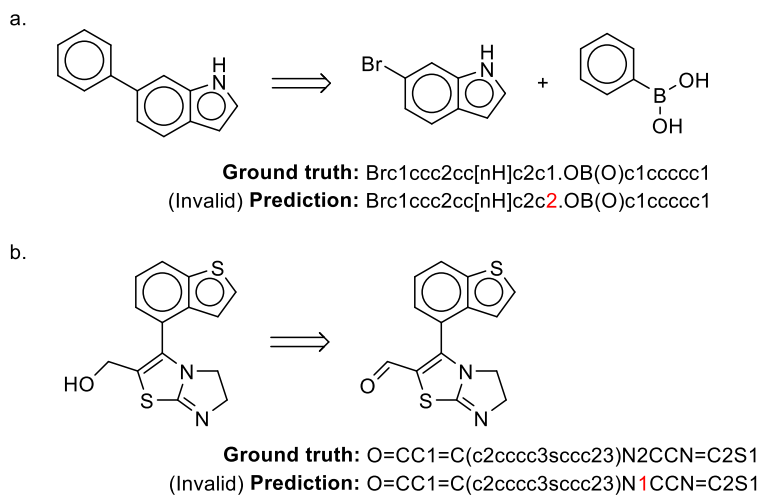

*Figure 6: Examples of predicted reactant SMILES that are grammatically invalid*

2. The predicted reactant SMILES is grammatically valid, but the overall reaction is not chemically plausible. This is a typical error in which the predicted reactant set cannot react in the specified reaction type to produce the target molecule. Many of these errors can also be attributed to the fragile SMILES text representation because small alterations in the SMILES can result in very large differences in the resulting molecule. Figure 7 shows examples of this type of error.

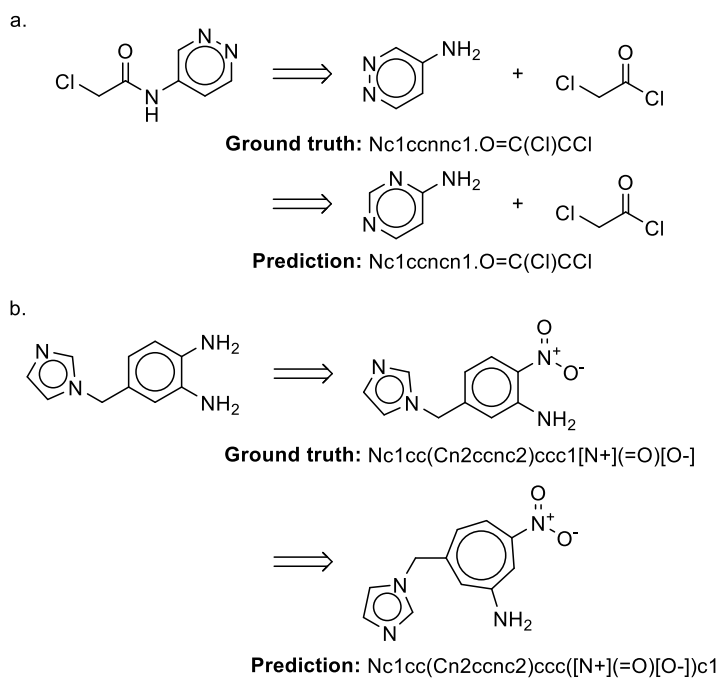

*Figure 7: Examples of predicted reactant SMILES that are grammatically valid, but the overall reaction is chemically implausible*

3. The predicted reactant SMILES is grammatically valid and the overall reaction is chemically plausible. Although the predicted reactant set does not match the ground truth reactant set, the predicted reactant set is likely to react in the specified reaction type to produce the target molecule.

One reason for this type of error is the possibility of multiple possible functional groups combinations or reactants that can react in the same reaction type to form the target molecule. Another reason is the presence of multiple reaction sites in the target molecule that can be disconnected retrosynthetically, so multiple possible reactant sets are chemically plausible. Figure 8 shows examples of this type of error.

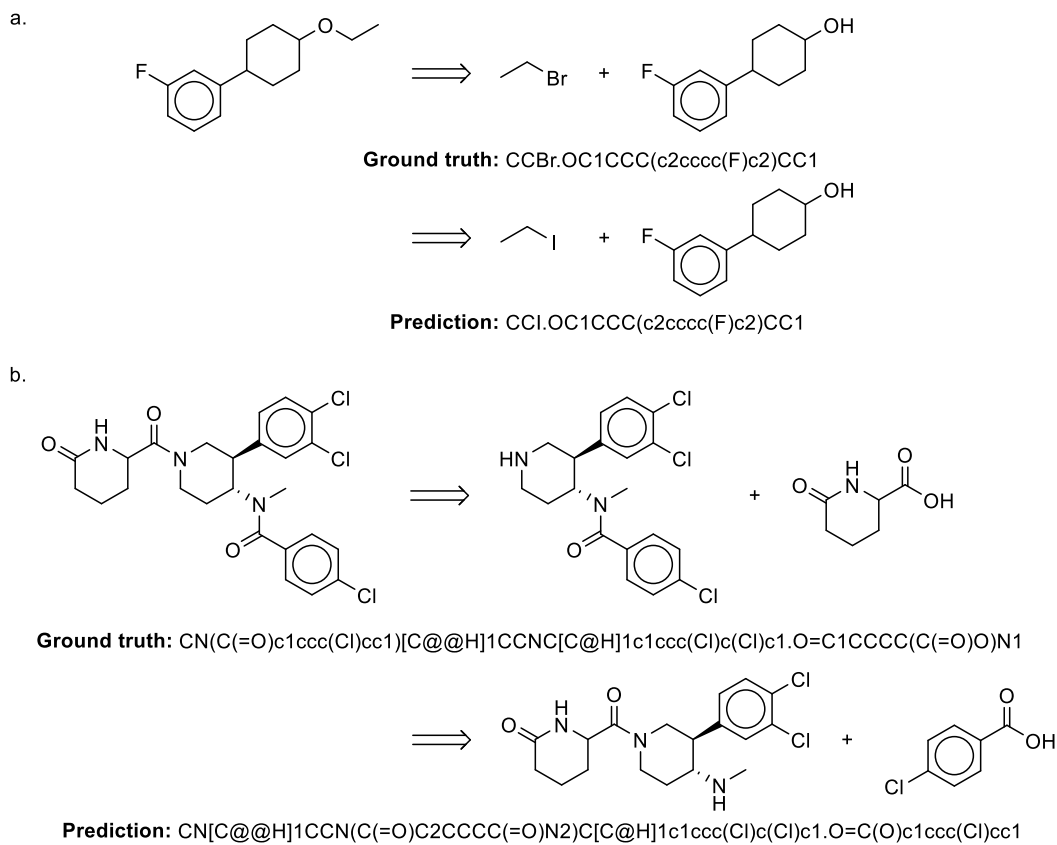

Figure 8: Examples of predicted reactant SMILES that are grammatically valid and the overall reaction is chemically plausible

*Ranking of the seq2seq model predictions*

The ranked predictions from the beam search decoding procedure of seq2seq model correspond well to chemical reactivity. For the top-10 predictions with both the baseline and seq2seq models, Figure 9 depicts a histogram that shows the counts of the highest rank that is assigned to the prediction which matches the ground truth for each example in the test dataset. Overall, the higher the rank of the prediction, the more likely that the prediction corresponds to the ground truth. The distribution for the seq2seq model is much more skewed towards the highest ranks compared to the baseline model, which naively ranks by the number of occurrences of the rules that was observed in the training dataset.

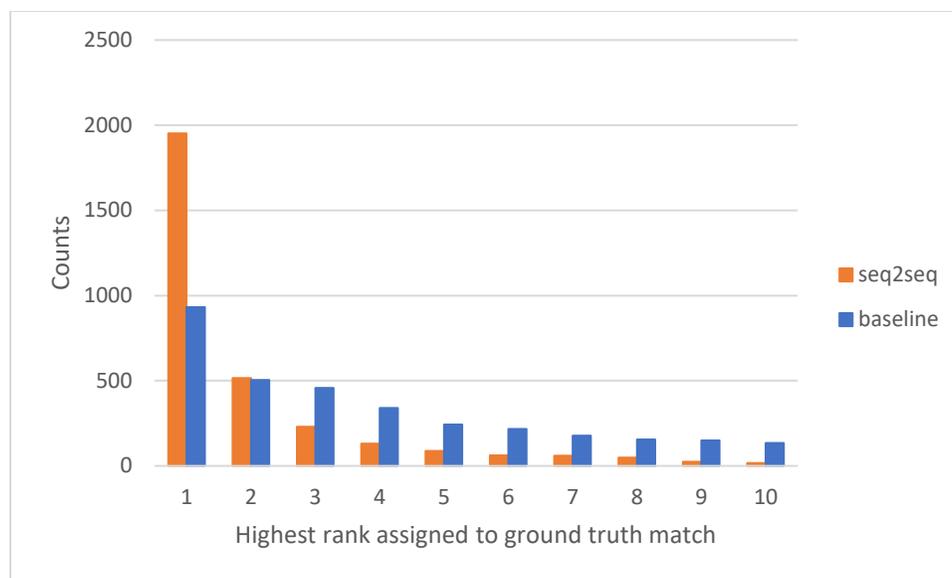

*Figure 9: Histogram of the highest rank assigned to the ground truth match in the top-10 predictions of the seq2seq and baseline models for each example. Note that the relative total counts across all the ranks for the seq2seq and baseline models is proportional to their relative top-10 accuracies shown in Table 2.*

**Discussion**

The seq2seq model performs comparably to the rule-based baseline model on the processed patent dataset, although the models perform differently for certain reaction types. Importantly, the seq2seq model has some significant advantages compared to the baseline model, and by extension to the deep learning based approaches that combine a rule-based expert system with a NN model for candidate ranking.

Firstly, the seq2seq model can be trained in a fully end-to-end manner directly from the training dataset. The seq2seq model both implicitly learns the chemical rules, and performs candidate ranking via the beam search decoding procedure. Conversely, the typical deep learning approach to reaction prediction combines a rule-based expert system component with a NN model component for candidate ranking, where the individual components need to be independently set up and trained. Also, any rule-based expert system that automatically extracts reaction rules from the reaction dataset depend heavily on accurate atom-mapping to describe the correspondence between the reactant and product atoms, which is itself a non-trivial problem.[55] The seq2seq model does not require atom-mapped reaction examples for training.

Secondly, the seq2seq model scales better to larger training datasets. The efficiency of rule-based expert systems depends on the number of rules in the knowledge base, which is an issue because the size of the knowledge base generally increases as the size of the training dataset increases. For the baseline model, as well as any deep learning approaches that use a rule-based expert system and a NN component to rank the likelihood of each predicted molecular species for a given example,[37] the inference cost directly depends on the size of the knowledge base since every rule in the knowledge base must be exhaustively applied. On the other hand, the inference cost of a particular seq2seq model is independent of the size of the training dataset and depends primarily on the width of the beam search decoding procedure. For deep learning approaches that use a rule-based expert system and a NN component to rank the applicability of each rule in the knowledge base for a given example,[35,36] a previous study showed that the classification accuracy decreases as the size of the knowledge base increases for a specific training dataset size, likely because the number of reaction rules that needs to be ranked in the multiclass classification problem also increases.[35]

Overall, as training dataset size increases, the increasing NN accuracy from more training examples is partially offset by the negative effect on accuracy from the increased number of reaction rules to classify. In order to reduce the number of rules in the knowledge base, previous studies[35,37] removed rare reaction rules that occurred fewer times than a specified threshold. A side effect of this is a reduced rule coverage over the reaction examples, especially for rare reactions types.

Thirdly, the seq2seq model incorporates information about the global molecular environment since the model learns from the complete SMILES for both the reactants and the target molecule in each reaction example. Also, the complete input target molecule SMILES is used to make predictions. However, the baseline model focuses only on the local molecular environment because the automatically extracted reaction rules in the knowledge base only incorporate the immediately neighboring atoms around the reaction centers. This results in an important issue with rule-based expert systems when performing the retrosynthetic reaction prediction task. In particular, the reaction examples in the dataset are processed into the form depicted in Figure 2, where each reaction consists of a single target molecule and one or more reactants. In all cases, every atom in the target molecule is atom-mapped and can be linked to reactant atoms. Unfortunately, the inverse is not true because for most reaction classes, not all atoms in the reactants are atom-mapped. These unmapped reactant atoms are the leaving groups which are not incorporated into the target molecule structure in the forward reaction. The issue arises when the leaving groups are large, which occurs commonly in reaction classes 5 (protections) and 6 (deprotections). If very general reaction rules are extracted that only contain the immediate neighborhood of the reaction centers, they will not fully capture the identities of the leaving groups. As a result, for these reaction classes that involve large leaving groups, the rules do not have sufficient information to reproduce the reactants from the target molecule. One solution would be to extract more specific rules that contains a larger neighborhood around the reaction centers in order to capture the identities of the leaving groups. However, the disadvantage with more specific rules is that they are less generalizable to new examples in the test dataset. Ultimately, there is a tradeoff between defining very general reaction rules and defining very specific reaction rules, which is challenging without manual intervention. Furthermore, a direct benefit of the seq2seq model incorporating the global molecular environment is that the model naturally accounts for stereochemistry. On the other hand, in order for deep learning approaches that combine a rule-based expert system with a NN model component to account for stereochemistry, both the reaction rules in the knowledge base and the descriptors that are used in the NN model must incorporate stereochemistry. Existing algorithms that automatically extract reaction rules do not fully address the issue of stereochemistry.

**Conclusion**

Overall, we have created a data driven neural sequence-to-sequence model to solve the retrosynthetic reaction prediction task, which is a critical task in computational retrosynthetic analysis. While the current implementation of the seq2seq model performs comparably to the rule-based expert system baseline, the seq2seq model has fundamental advantages over rule-based expert systems and over any deep learning approach that depends on a rule-based expert system component. The seq2seq model can be trained in an end-to-end manner, scales more efficiently to larger datasets, and naturally incorporates the global molecular environments of the reaction species. We believe that there exists significant room for improvement over the current relatively unoptimized seq2seq model, so further work is likely to lead to greater prediction accuracies. In following work, we will seek to increase the accuracy of the seq2seq model by exploring architectural variants and by exploring new datasets. Furthermore, an interesting extension to the seq2seq models would be to use one-shot techniques[56–58] to allow our retrosynthetic reaction prediction models to make reasonable predictions for reaction classes where only a few example reactions are available for training.

We believe that the approach and model architecture described in this work is an important early step towards solving the computational retrosynthetic analysis problem. Although the chemical complexity of the reactions and molecules explored in this work are intentionally simple to facilitate analysis and thus are quite far away from what is typically faced by mainstream synthetic chemists today, we strongly believe that the future evolution of this approach could bridge this gap and result in tools that can become broadly useful for the expert chemist and also more broadly for those less skilled in the science.

**Acknowledgements**

We thank Franklin Lee for critical reading and feedback on the manuscript.

B.L. is supported by the NIH (U19 AI109662). B.R. is supported by the Fannie and John Hertz Foundation. S.H. is supported by a Postdoctoral Fellowship, PF-15-007-01-CDD from the American Cancer Society.

The Pande Group is broadly supported by grants from the NIH (R01 GM062868 and U19 AI109662) as well as gift funds and contributions from Folding@home donors. This research was also supported by the National Science Foundation (P.A.W.: CHE1265956).

We acknowledge the generous support of Dr. Anders G. Frøseth for our work on machine learning.

**Declarations**

VSP is a consultant & SAB member of Schrodinger, LLC and Globavir, sits on the Board of Directors of Apeel Inc, Freenome Inc, Omada Health, Patient Ping, Rigetti Computing, and is a General Partner at Andreessen Horowitz

**Supporting information**

*Table S1: Key hyperparameters of the seq2seq model*

| General | |
|---|---|
| Batch size | 32 |
| Optimizer | Adam |
| Learning rate | 0.0001 |
| Max gradient norm | 5.0 |
| Max sequence length | 140 |
| Attention dim | 512 |
| | |
| ***Bidirectional LSTM encoder*** | |
| # layers | 2 |
| Embedding dim | 512 |
| Dropout (keep) | 0.8 |
| | |
| ***LSTM decoder*** | |
| # layers | 4 |
| Embedding dim | 512 |
| Dropout (keep) | 0.8 |
| Max decode length | 140 |